\documentclass{article}
\usepackage{spconf,amsmath,epsfig}
\usepackage{multirow,subfigure}
\usepackage{float}
\usepackage{caption}
\usepackage{cite}
\usepackage[most]{tcolorbox}
\usepackage{graphicx}
\usepackage{subcaption}
\usepackage{url}
\usepackage{xcolor}
\usepackage{booktabs}

\definecolor{cl1}{rgb}{0.93, 0.49, 0.19}
\newcommand{\CLA}[1]{\textcolor{cl1}{#1}}
\definecolor{cl2}{rgb}{0, 0.6, 0}
\newcommand{\CLB}[1]{\textcolor{cl2}{#1}}
\definecolor{cl3}{rgb}{1,0,0}
\newcommand{\CLC}[1]{\textcolor{cl3}{#1}}

\let\OLDthebibliography\thebibliography
\renewcommand\thebibliography[1]{
  \OLDthebibliography{#1}
  \setlength{\parskip}{0pt}
  \setlength{\itemsep}{0pt plus 0.3ex}
}

\begin{document}\sloppy

\def\x{{\mathbf x}}
\def\L{{\cal L}}

\title{Q-Refine: A Perceptual Quality Refiner for AI-Generated Image}
%
\names{Chunyi Li\textsuperscript{1,2} \qquad Haoning Wu\textsuperscript{3} \qquad Zicheng Zhang\textsuperscript{1} \qquad Hongkun Hao\textsuperscript{1} \qquad Kaiwei Zhang\textsuperscript{1}}{Lei Bai\textsuperscript{2} \qquad Xiaohong Liu\textsuperscript{1} \qquad Xiongkuo Min\textsuperscript{1} \qquad Weisi Lin\textsuperscript{3} \qquad Guangtao Zhai\textsuperscript{1}}
\address{Shanghai Jiao Tong University\textsuperscript{1}, Shanghai AI Lab\textsuperscript{2}, Nanyang Technological University\textsuperscript{3}}
{
\twocolumn[{
\renewcommand\twocolumn[1][]{#1}
\maketitle

\begin{center}
    \vspace{-4mm}
    \includegraphics[width = \textwidth]{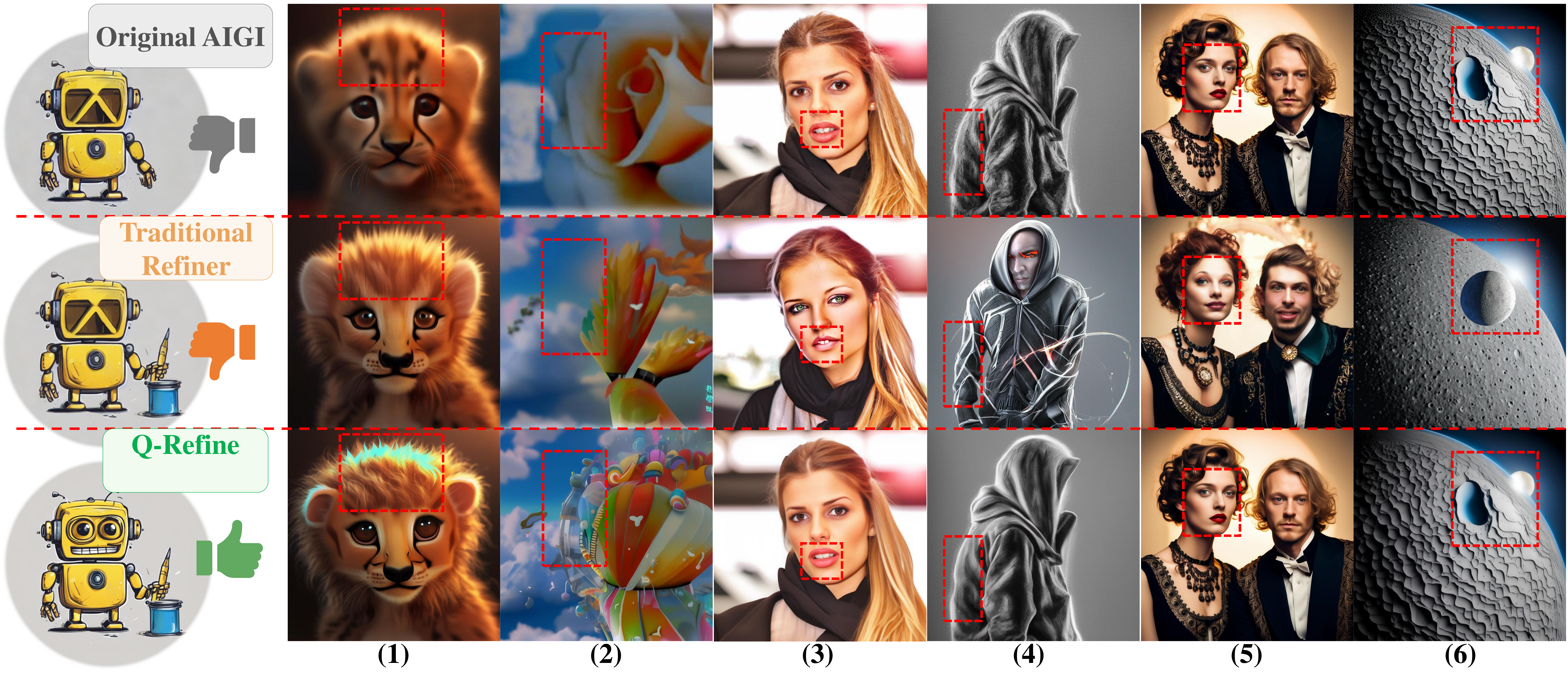}
    \captionof{figure}{The original AIGIs from AGIQA-3K\cite{database:agiqa-3k}, optimized by \textbf{\CLA{Traditional Refiners}} and \textbf{\CLB{Q-Refine}} we proposed. 
    As a quality-aware metric, the Q-Refine can add details on the blurred part, to better optimize low-quality regions of (1)(2); improve clarity in medium-quality regions of (3)(4) without changing the whole image; and avoid degrading the high-quality regions of (5)(6).
    }
    \label{fig:spotlight}
\end{center}}]
}

\begin{abstract}
With the rapid evolution of the Text-to-Image (T2I) model in recent years, their unsatisfactory generation result has become a challenge. However, uniformly refining AI-Generated Images (AIGIs) of different qualities not only limited optimization capabilities for low-quality AIGIs but also brought negative optimization to high-quality AIGIs. To address this issue, a quality-award refiner named Q-Refine\footnote{The code will be released on https://github.com/Q-Future/Q-Refine} is proposed. Based on the preference of the Human Visual System (HVS), Q-Refine uses the Image Quality Assessment (IQA) metric to guide the refining process for the first time, and modify images of different qualities through three adaptive pipelines. Experimental shows that for mainstream T2I models, Q-Refine can perform effective optimization to AIGIs of different qualities. It can be a general refiner to optimize AIGIs from both fidelity and aesthetic quality levels, thus expanding the application of the T2I generation models.

\end{abstract}
\begin{keywords}
AI-Generated Content, Image Quality Assessment, Image Restoration
\end{keywords}

\section{Introduction}
\label{sec:intro}

AI-Generated Content (AIGC) refers to the creation of content, such as images, videos, and music, using AI algorithms \cite{database:agiqa-3k}. Since vision is the dominant way for humans to perceive the external world, AI-Generated Images (AIGIs) \cite{review:texttoimage} have become one of the most representative forms of AIGC. The development of Text-to-Image (T2I) models is a crucial step in the advancement of AIGIs, as it allows for the creation of high-quality images that can be used in a variety of applications\cite{review:3model}, including advertising, entertainment, and even scientific research. The importance of AIGI in today's internet cannot be overstated, as it has the potential to revolutionize the way we consume and interact with visual content. 

With the rapid technological evolution of T2I generation techniques, there have been at least 20 representative T2I models coexisting up to 2023, whose generation quality varies widely\cite{database:agiqa-3k}. Coupled with confusing prompt input, unreasonable hyper-parameter settings, and insufficient iteration epochs, the quality of today's AIGIs is still not satisfying.

\begin{figure*}[t]
    \centering
    \includegraphics[width = 0.9\linewidth]{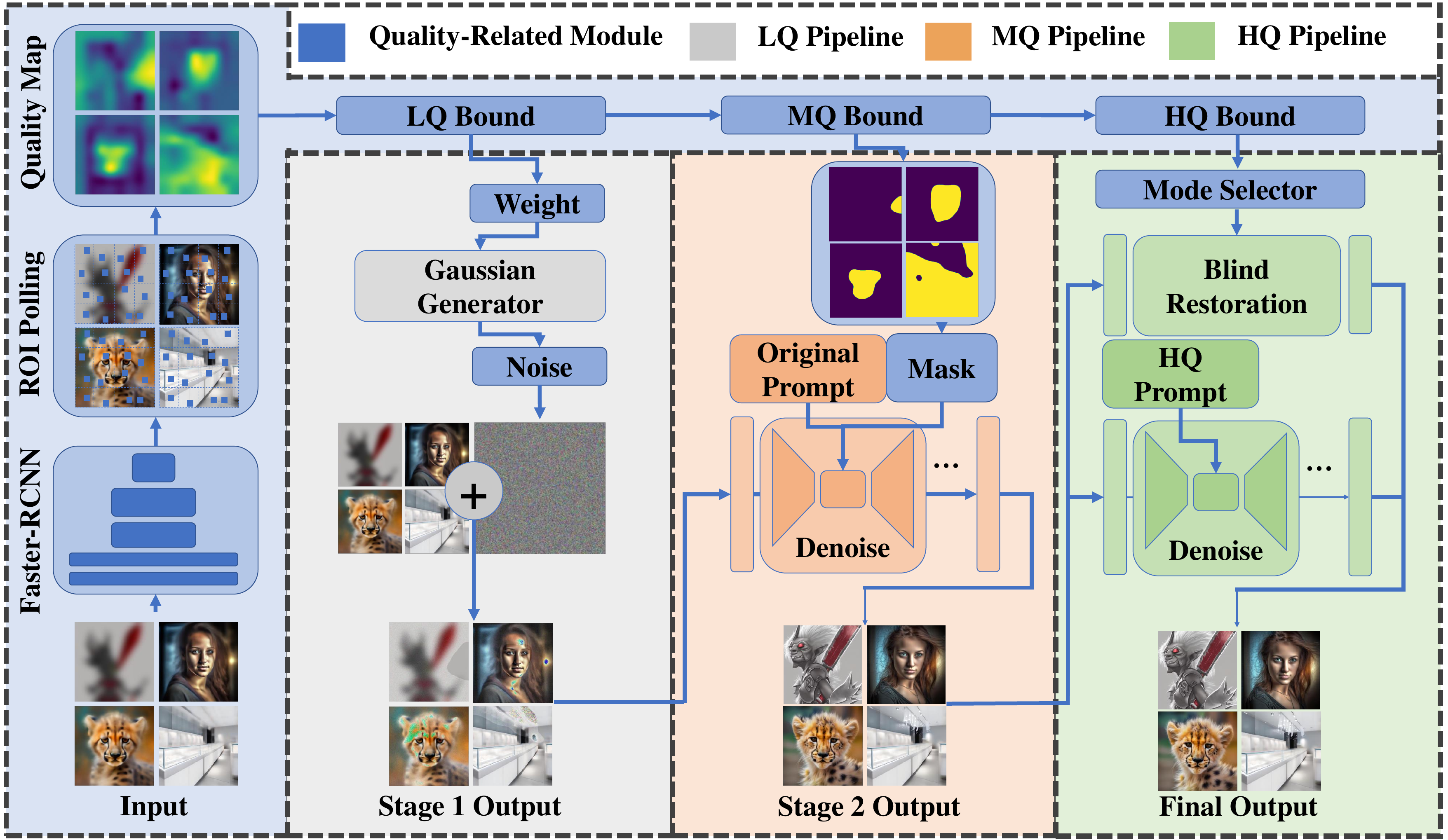}
    \caption{Framework of Q-Refine, including a quality pre-prossess module, and three refining pipelines for low/medium/high quality (LQ/MQ/HQ) regions. The refining mechanisms for each pipeline are inspired by the predicted quality.}
    \label{fig:framework}
\end{figure*}

Considering the wide application of AIGIs, their quality needs to be further optimized. However, this task is extremely challenging as shown in Fig. \ref{fig:spotlight}. Firstly, positive optimization is difficult to achieve for \textbf{Low-Quality} (LQ) regions. If their quality falls into a local optimum, they won't be modified as a global optimum; secondly, local negative optimization is a hidden danger of \textbf{Medium-Quality} (MQ) regions. Since the quality distribution of images varies, refiners need to change only the LQ/MQ without affecting other regions; finally, global negative optimization is common in \textbf{High-Quality} (HQ) regions. Since the performance of refiners has a certain limit, blindly modifying an already high-quality image can easily lead to a decrease in quality.

\section{Related Work and Contributions}
\label{sec:relate}

Existing AIGI quality refiners are mainly divided into two types. The most commonly used method is to treat AIGI as a Natural Sense Image (NSI) and use a large-scale neural network for Image Restoration ~\cite{metric:dasr,metric:diffbir,metric:rfdn}; the other is to use the prompt as guidance, then put the AIGI back into a generative model for several epochs ~\cite{gen:SD,gen:XL}. However, both refiners ignore image quality. Using the same pipeline for LQ/MQ/HQ will lead to insufficient enhancement in the LQ regions and negative optimization in the HQ regions, essentially bringing all images to the MQ level as Fig. \ref{fig:spotlight} shows. 

Therefore, the quality of AIGIs needs to be computed in advance as refining guidance. However, Image Quality Assessment (IQA) \cite{iqa:stablevqa,add:vdpve} and Refiner cannot be directly combined. Existing IQA works ~\cite{my:aspect-qoe,my:cartoon,my:xgc-vqa} usually consider the overall quality of the image, instead of a quality map, making it difficult for the refiner to implement local optimization.

To enhance positive while avoiding negative optimization, we found a way to combine IQA with refiners named Q-Refine, the first quality-aware refiner for AIGIs based on the preference of the Human Visual System (HVS) with the following contribution:
($i$) We introduce the IQA map to guide the AIGI refining for the first time. A new paradigm for AIGI restoration, namely using quality-inspired refining is proposed.
($ii$) We establish three refining pipelines that are suitable for LQ/MQ/HQ regions respectively. Each pipeline can self-adaptively determine the executing intensity according to the predicted quality.
($iii$) We extensively conduct comparative experiments between existing refiners and Q-Refine on mainstream AIGI quality databases. The result proved the strong versatility of Q-Refine.

\section{Proposed Method}
\label{sec:method}

\subsection{Framework}
\label{sec:framework}

Since perceptual quality has been widely recognized as a decisive role for Generative AI ~\cite{q-bench, q-instruct,q-boost},
Q-Refine is designed to refine AIGIs with separated pipelines according to the quality. Our framework is shown in Fig. \ref{fig:framework} with an IQA module to predict a quality map and three pipelines include: (1) Gaussian Noise: encouraging changing the LQ region by adding noise; (2) Mask Inpainting: generating a mask from the quality map to reserve HQ region; (3) Global Enhancement: setting an enhancement threshold to fine-tune the final output.

\subsection{IQA Module}
\label{sec:iqa}

Splitting the image into patches \cite{his:advancing}, evaluating them separately \cite{his:gms}, and then combining them is a commonly used \cite{iqa:paq} IQA pipeline in recent years. It can evaluate the overall quality while providing a rough quality map through patches. By dividing an AIGI into $n\times n$, a patch $P$ with index $(i,j)\in [0,n-1]$ has:
\begin{equation}
    {P_{(i,j)}} = {\rm CNN} (I_{(\frac{i}{n}h:\frac{{i + 1}}{n}h,\frac{j}{n}w:\frac{{j + 1}}{n}w)})
\end{equation}
where $(h,w)$ are the height/width of the input image $I$. Since extracting the quality map requires a network sensitive for both global regression and local perception, the dual-task structure for image classification/detection, namely Faster-RCNN\cite{iqa:fasterrcnn}, is utilized as our ${\rm CNN}$ model backbone. For local quality ${Q_{(i,j)}}$, referring to previous quality map extractor\cite{iqa:paq}, we use the largest value in each patch as its quality score, to obtain a $n\times n$ quality map $Q$. However, for global quality $q$, to avoid excessive complexity affecting the subsequent three refining pipelines, we abandoned all global extractors and directly averaged the patch scores as:
\begin{equation}
    \left\{ \begin{array}{l}
{Q_{(i,j)}} = {\rm RoIPool}({P_{(i,j)}})\\
q = {\rm Avg}({Q_{(i,j)}})
\end{array} \right.
\end{equation}
where ${\rm Avg}$ and ${\rm RoIPool}$ are the average and average-max-pooling layers. The global quality/quality map will guide refining pipelines.

\subsection{Stage 1 Pipeline: Gaussian Noise}
\label{sec:LQ}

\begin{figure}[t]
    \centering
    \includegraphics[width = 0.9\linewidth]{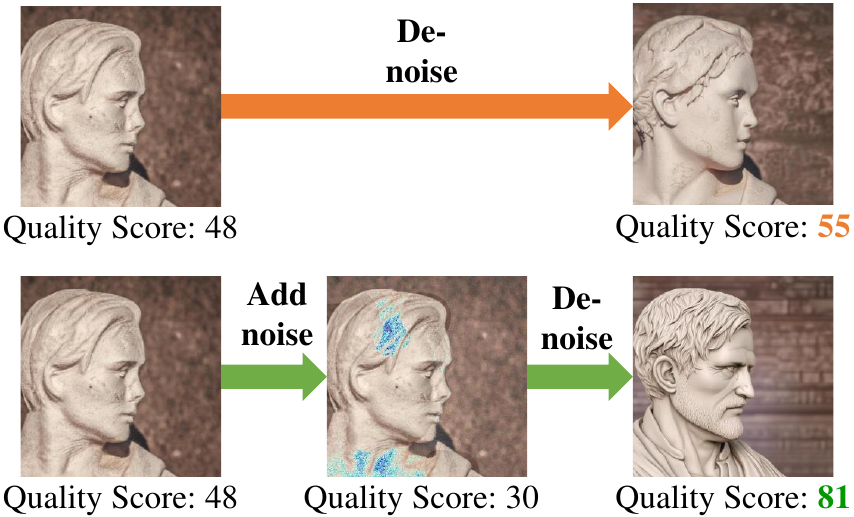}
    \caption{The refining result by \textbf{\CLA{only denoise}} / \textbf{\CLB{add noise + denoise}} from SDXL\cite{gen:XL}. 
    Adding noise reduces quality \cite{iqa:paq}, but it lays the foundation for global optimality before denoising.}
    \label{fig:lq}
\end{figure}
Existing T2I generation models cannot always ensure a HQ result, even the most advanced model \cite{gen:pixart} may occasionally generate blurry images. Such a problem may be due to the initial few denoising steps, causing the image to fall into a local optimum. In such cases, the model will stubbornly retain some LQ regions, causing the image to remain unchanged even after iterating hundreds of epochs. To solve this problem, such LQ regions should rewind to previous steps, to trigger the model's denoising mechanism. Since Sec. \ref{sec:iqa} provides a quality map, the LQ region can be identified and then modified. As the starting noise image before denoising, we superimpose Gaussian noise in the LQ region to obtain the first stage output $I_{s1}$:
\begin{equation}
\left\{ \begin{array}{l}
W = \max ({B_{LQ}} - Q,0)\\
I_{s1} = W\mathcal{G}_{(h,w)} + (1 - W)I
\end{array} \right.
\end{equation}
where the noise weight map $W$ is determined by LQ bound $B_{LQ}$, a region with lower quality has higher weight while quality larger than $B_{LQ}$ leads to zero weight. The size of Gaussian noise $\mathcal{G}$ is $(h,w)$. As Fig. \ref{fig:lq} shows, though the noise from the stage 1 pipeline may temporarily reduce the image quality, it can help the following two pipelines to change the LQ region. By refining the final output, it can move the local quality optimum toward the global optimum.

\subsection{Stage 2 Pipeline: Mask Inpainting}
\label{sec:MQ}

\begin{figure}[t]
    \centering
    \subfigure[Patch quality map]{\includegraphics[width = 0.23\textwidth]{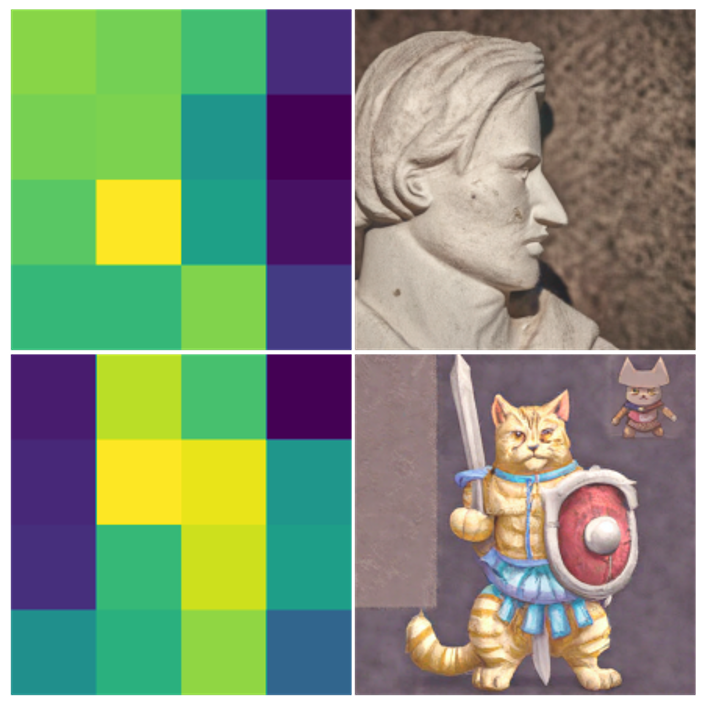}
    }
    \subfigure[Flattened quality map]{\includegraphics[width = 0.23\textwidth]{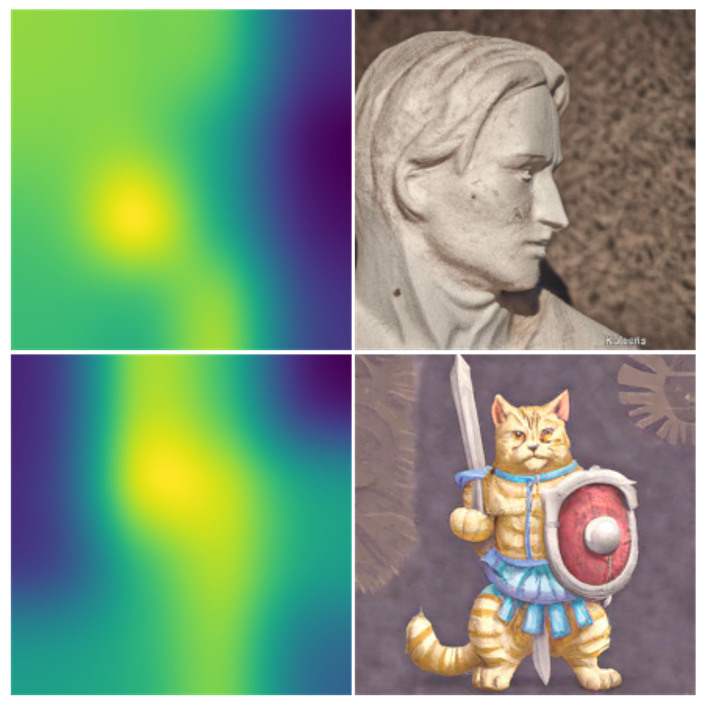}
    }
    \caption{Using original patch quality map / flattened map to guide the inpainting. (a) suffers from block effects and unexpected artifacts while (b) has a smooth and natural result.}
    \label{fig:mq}
\end{figure}

Since different regions of images have different quality, this pipeline aims to retain HQ and modify other regions. This operation can be completed through the inpainting method, by taking LQ regions as a mask. However, as the edges between patches are un-discontinuous, directly using the quality map with $n\times n$ patches to generate this mask will cause some unsatisfying results like Fig. \ref{fig:mq} shows. First, a discontinuous quality map may require the inpainting model to retain a certain patch and modify adjacent patches. The result will have obvious block effects at the edge of the patches. Second, the inpainting model tends to redraw the main object with a regular rectangle mask. Though we only want some detail on a plain background, it will generate unexpected main objects here instead. Thus the patch quality map $Q$ needs to be flattened before inpainting. Considering smoothness is our first requirement, we use the smoothest interpolation method Bi-Cubic\cite{mq:bicubic}, to convolve each pixel with 16 adjacent pixels:
\begin{equation}
Q_{(x,y)} =  \sum\limits_{r, c =  - 1}^2 {{Q_{(\left\lfloor {x\frac{n}{h}} \right\rfloor  + r,\left\lfloor {y\frac{n}{w}} \right\rfloor  + c)}}{{Cub}_{(r - x,c - y)}}}  
\end{equation}
where pixel $(\left\lfloor {x\frac{n}{h}} \right\rfloor,\left\lfloor {y\frac{n}{w}} \right\rfloor)$ from the original quality map is the corresponding pixel $(x,y)$ from the flattened map and $Cub$ stands for the Bi-Cubic\cite{mq:bicubic} matrix. From this, the probability density function $\bf{z}$ of each step is:
\begin{equation}
{\bf{z}} = {\rm{QKV}}(prompt,mask = \{Q - B_{MQ}\})
\end{equation}
where we set quality region below the threshold $B_{MQ}$ as mask. QKV stands for multi-head attention, which depends on the input $prompt$ and $mask$. Set the starting point of denoising to $x_0=I_{s1}$, we have the second stage output $I_{s2}$:
\begin{equation}
I_{s2}=x_m={{\cal D}_m}({x_{m - 1}})= {{\cal D}_m}({\cal D}_{m-1}\cdots{{\cal D}_1}({I_{s1}}))
\label{equ:diffusion}
\end{equation}
where $\mathcal{D}_m$ represents the diffusion operation at the $m$-th iteration and $x$ stands for this intermediate state. From this, we used masks to modify the LQ/MQ region through the smoothed quality map without affecting the HQ region.

\begin{figure}[t]
    \centering
    \includegraphics[width = \linewidth]{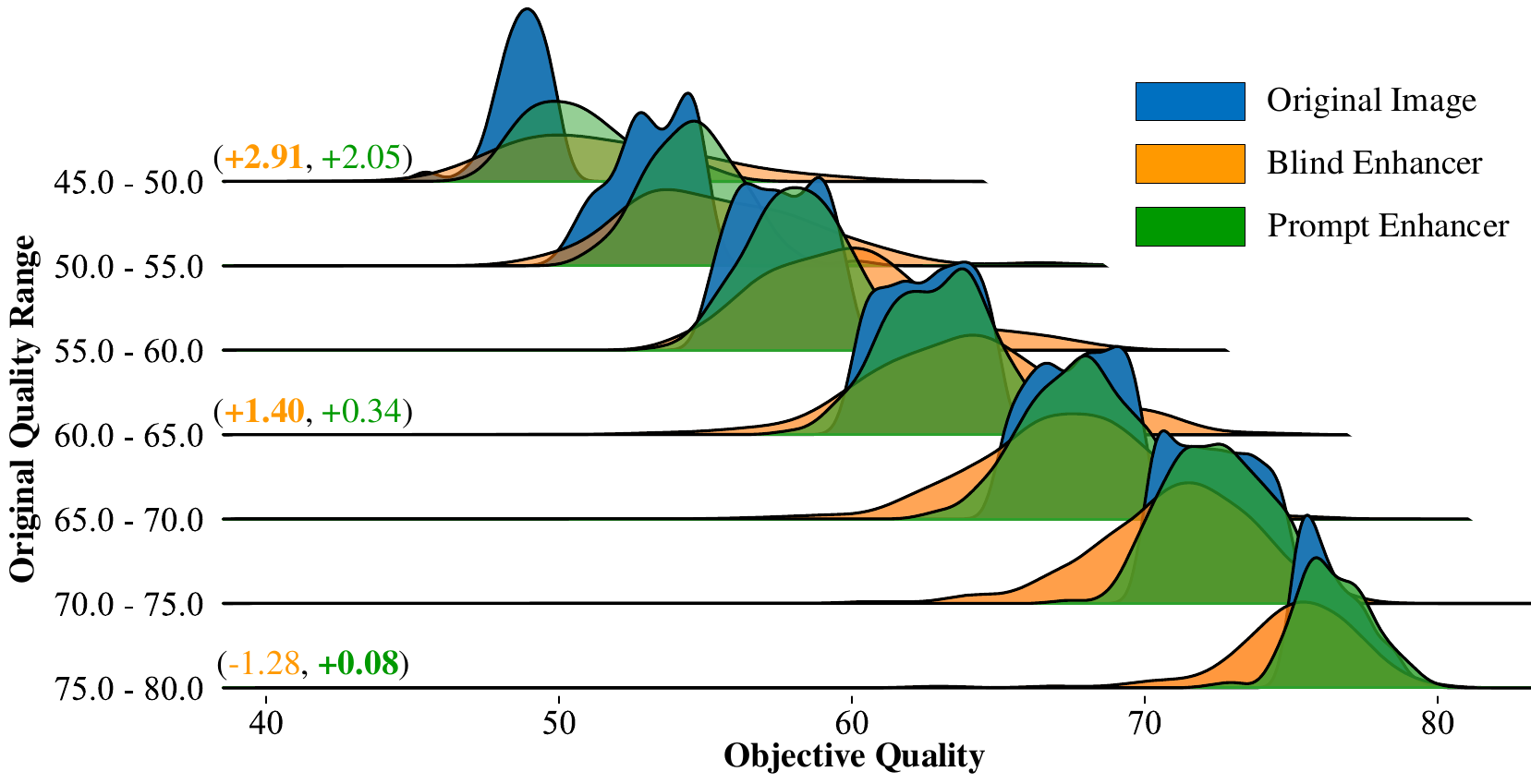}
    \caption{Using \textbf{\CLA{blind enhancer}} or \textbf{\CLB{prompt-guided enhancer}} to refine images in different quality groups in AGIQA-3K\cite{database:agiqa-3k}. Blind enhancer shows better refining results for LQ groups but causes negative optimization for HQ groups.}
    \label{fig:hq}
\end{figure}

\subsection{Stage 3 Pipeline: Global Enhancement}
\label{sec:HQ}

After local inpainting, to further improve the image quality, this pipeline fine-tunes the image's low-level attributes, rather than adding new objects. Low-level quality enhancers include the following two types. One is the traditional image super-resolution/restoration method, which ignores the prompt as a blind enhancer, using prior knowledge from NSIs to achieve image-to-image reconstruction. The other is the generative model, which uses the original prompt to guide the diffusion for several steps, namely prompt-guided enhancer. The SOTAs of the two enhancers are DiffBIR\cite{metric:diffbir} and SDXL\cite{gen:XL}, and the refining results are in Fig. \ref{fig:hq}. Considering the blind enhancer is suitable for LQ, but performs worse than the prompt-guided enhancer on HQ, we implement the enhancer based on global quality, with the final output $I_{f}$:

\begin{equation}
I_{f}= \{{\rm E}_B,{\rm E}_P \Vert q<B_{HQ}\}(I_{s2})
\end{equation}
where ${\rm E}_B$ stands for a blind enhancer while ${\rm E}_P$ performs a similar mechanism as (\ref{equ:diffusion}), but in smaller hyper-parameter strength (to avoid negative optimization for HQ) without a mask. The HQ bound $B_{HQ}$ determines such selection. Meanwhile, considering some positive words \cite{hq:prompt} will significantly improve the generation quality, we combine these words with the original prompt as the input of ${\rm E}_P$. Therefore, regardless of whether the input belongs to LQ/MQ/HQ, our model can refine its quality by providing an HQ result.

\section{Expriment}
\label{sec:exp}

\begin{table*}[tbph]
\centering
\caption{Refined result of AGIQA-3K \cite{database:agiqa-3k} database and five subsets from different generators. The refined results with the best quality are noted in \textbf{\CLC{red}}. The refined quality below the original data is noted in \underline{underline}.}
\label{tab:agiqa}
\subtable{
\begin{tabular}{l|ccc|ccc|ccc}
\toprule
             \multirow{2}{*}{Refiner} & \multicolumn{3}{c|}{Mean}                                                                                                & \multicolumn{3}{c|}{GLIDE\cite{gen:GLIDE}}                                                                                               & \multicolumn{3}{c}{SDXL\cite{gen:XL}}                                                                                              \\ \cline{2-10} 
             & Overall$\uparrow$                               & Aesthetic$\uparrow$                              & Fidelity$\downarrow$                              & Overall$\uparrow$                               & Aesthetic$\uparrow$                              & Fidelity$\downarrow$                              & Overall$\uparrow$                                & Aesthetic$\uparrow$                              & Fidelity$\downarrow$                              \\ \hline
Original     & 0.5710                                 & 0.4890                                 & 38.975                                 & 0.2901                                 & 0.2895                                 & 71.331                                 & 0.7559                                 & 0.6173                                 & 24.816                                 \\ 
DASR\cite{metric:dasr}         & {\underline{0.4987}} & 0.5507                                 & {\underline{45.252}} & {\underline{0.2384}} & 0.3007                                 & 63.922                                 & {\underline{0.7298}} & 0.7011                                 & 22.728                                 \\
DiffBIR\cite{metric:diffbir}      & 0.5829                                 & 0.5935                                 & 35.049                                 & 0.4104                                 & 0.3982                                 & 60.728                                 & {\underline{0.7400}} & {\color[HTML]{FF0000} \textbf{0.7273}} & {\underline{26.309}} \\
RFDN\cite{metric:rfdn}         & {\underline{0.5704}} & {\underline{0.4885}} & 38.831                                 & {\underline{0.2900}} & {\underline{0.2886}} & 71.178                                 & {\underline{0.7532}} & {\underline{0.6164}} & 24.522                                 \\ 
SD1.5\cite{gen:SD} & 0.6461                                 & 0.5359                                 & {\underline{39.649}} & 0.5852                                 & 0.4749                                 & 67.669                                 & {\underline{0.6917}} & {\underline{0.5632}} & {\underline{29.996}} \\
SDXL\cite{gen:XL}  & 0.6489                                 & 0.5418                                 & 32.999                                 & 0.5609                                 & 0.4416                                 & 52.711                                 & {\underline{0.7111}} & {\underline{0.5842}} & 24.589                                 \\ 
Q-Refine     & {\color[HTML]{FF0000} \textbf{0.7232}} & {\color[HTML]{FF0000} \textbf{0.6021}} & {\color[HTML]{FF0000} \textbf{22.463}} & {\color[HTML]{FF0000} \textbf{0.6333}} & {\color[HTML]{FF0000} \textbf{0.4986}} & {\color[HTML]{FF0000} \textbf{31.722}} & {\color[HTML]{FF0000} \textbf{0.8007}} & 0.6640                                 & {\color[HTML]{FF0000} \textbf{18.145}} \\
\bottomrule
\end{tabular}
}
\subtable{
\begin{tabular}{l|ccc|ccc|ccc}
\toprule
             \multirow{2}{*}{Refiner} & \multicolumn{3}{c|}{DALLE2\cite{gen:DALLE2}}                                                                                              & \multicolumn{3}{c|}{MidJourney\cite{gen:MJ}}                                                                                           & \multicolumn{3}{c}{SD1.5\cite{gen:SD}}                                                                                                 \\ \cline{2-10} 
             & Overall$\uparrow$                               & Aesthetic$\uparrow$                              & Fidelity$\downarrow$                              & Overall$\uparrow$                               & Aesthetic$\uparrow$                              & Fidelity$\downarrow$                              & Overall$\uparrow$                                & Aesthetic$\uparrow$                              & Fidelity$\downarrow$                              \\ \hline
Original     & 0.6193                                 & 0.4884                                 & 29.264                                 & 0.5340                                 & 0.4751                                 & 42.6938                                 & 0.6555                                 & 0.5749                                 & 26.7706                                 \\
DASR\cite{metric:dasr}         & {\underline{0.5686}} & 0.5884                                 & {\underline{38.917}} & {\underline{0.4521}} & 0.5562                                 & 39.6575                                 & {\underline{0.5045}} & 0.6069                                 & {\underline{61.0347}}          \\
DiffBIR\cite{metric:diffbir}     & {\underline{0.5947}}          & 0.6118                                 & 27.658                                 & 0.5543                                 & 0.5706                                 & 31.3149                                 & {\underline{0.6153}} & {\color[HTML]{FF0000} \textbf{0.6598}} & {\underline{29.2356}} \\
RFDN\cite{metric:rfdn}        & {\underline{0.6191}} & {\underline{0.4875}} & 29.120                                 & {\underline{0.5337}} & 0.4755 & {\underline{42.6950}}          & 0.6561 & {\underline{0.5745}} & 26.6402                                 \\
SD1.5\cite{gen:SD} & 0.6543                                 & 0.5425                                 & {\underline{33.885}} & 0.6295                                 & 0.5305                                 & 39.8227                                 & 0.6696 & {\underline{0.5686}} & {\underline{26.8717}} \\
SDXL\cite{gen:XL}  & 0.6692                                 & 0.5654                                 & {\underline{29.789}}          & 0.6307                                 & 0.5359                                 & 34.2414                                 & 0.6726 & 0.5819 & 23.6660                                 \\
Q-Refine     & {\color[HTML]{FF0000} \textbf{0.7350}} & {\color[HTML]{FF0000} \textbf{0.6133}} & {\color[HTML]{FF0000} \textbf{20.763}} & {\color[HTML]{FF0000} \textbf{0.7384}} & {\color[HTML]{FF0000} \textbf{0.6097}} & {\color[HTML]{FF0000} \textbf{19.4677}} & {\color[HTML]{FF0000} \textbf{0.7084}} & 0.6249                                 & {\color[HTML]{FF0000} \textbf{22.2168}} \\
\bottomrule
\end{tabular}
}
\vspace{-8mm}
\end{table*}

\subsection{Expriment Settings}
\label{sec:set}

Our Q-Refine is validated on three AIGI quality databases, including AGIQA-3K, AGIQA-1K, and AIGCIQA ~\cite{database:agiqa-3k, database:agiqa-1k, database:aigciqa}. The quality of AIGIs before/after Q-Refine is compared to prove the general optimization level. Moreover, since AGIQA-3K\cite{database:agiqa-3k} includes five T2I models ~\cite{gen:SD,gen:XL,gen:GLIDE,gen:DALLE2,gen:MJ} with remarkable quality differences, their performances are listed respectively to prove Q-Refine's versatility on LQ/MQ/HQ regions. Besides the original image, the image quality generated by Q-Refine is compared with three latest image restoration refiners ~\cite{metric:dasr,metric:diffbir,metric:rfdn} and two representative generative refiners~\cite{gen:SD,gen:XL} as Sec. \ref{sec:relate} reviewed.

To measure the image quality, since FID\cite{quality:fid} is inconsistent with human subjective preferences, we use IQA methods to represent HVS's perceptual quality. The image quality consists of two different levels. Signal-fidelity characterizes low-level quality including factors like blur or noise, which is the traditional definition of image quality. Thus, we use the classic Brisque\cite{quality:Brisque} as its index. Aesthetic, however, represents high-level quality, which depends on the overall appeal and beauty of the image. Here we take the HyperIQA \cite{quality:HyperIQA} as the index since it best correlates human subjective preference on AIGIs. Moreover, for a more intuitive performance comparison, we also take CLIPIQA \cite{quality:CLIPIQA} as an overall quality indicator for both levels.

\subsection{Expriment Result and Discussion}
\label{sec:result}
\begin{table}[t]
\centering
\caption{Three AIGI quality databases \cite{database:agiqa-3k,database:agiqa-1k,database:aigciqa} before/after Q-Refine. The best result is noted in \textbf{\CLC{red}}.}
\label{tab:all}
\begin{tabular}{l|ccc}
\toprule
                   Databases & Overall$\uparrow$                                & Aesethic$\uparrow$                              & Fidelity$\downarrow$                               \\ \hline
AGIQA-3K\cite{database:agiqa-3k}           & 0.5710                                 & 0.4890                                 & 38.975                                 \\
AGIQA-3K + Q-Refine & {\color[HTML]{FF0000} \textbf{0.7232}} & {\color[HTML]{FF0000} \textbf{0.6021}} & {\color[HTML]{FF0000} \textbf{22.463}} \\ \hline
AGIQA-1K\cite{database:agiqa-1k}            & 0.6454                                 & 0.5896                                 & 42.288                                 \\
AGIQA-1K + Q-Refine & {\color[HTML]{FF0000} \textbf{0.7258}} & {\color[HTML]{FF0000} \textbf{0.6511}} & {\color[HTML]{FF0000} \textbf{27.767}} \\ \hline
AIGCIQA\cite{database:aigciqa}             & 0.5720                                 & 0.5213                                 & 31.443                                 \\
AIGCIQA + Q-Refine  & {\color[HTML]{FF0000} \textbf{0.6639}} & {\color[HTML]{FF0000} \textbf{0.6196}} & {\color[HTML]{FF0000} \textbf{23.365}} \\
\bottomrule
\end{tabular}
\end{table}

The experimental performance on the AGIQA-3K\cite{database:agiqa-3k} database and five subsets is shown in Table \ref{tab:agiqa}. In the general perspective, Q-Refine achieved the best aesthetic, fidelity, and overall quality. On a total of 18 indexes in six sets, Q-Refine \textbf{reached SOTA on 16} of them. It is worth mentioning that Q-Refine \textbf{never negatively optimized} any index that other Refiners never achieved.
From a detailed perspective, Q-refine has a satisfying performance on all subsets as we stated in our contributions. Firstly, for the worst quality GLIDE\cite{gen:GLIDE} model, the significant improvement of the three indexes proves that Q-Refine can effectively refine LQ. Secondly, for the strongest SDXL\cite{gen:XL} model, each index after Q-Refine does not drop like other methods certified the robustness on HQ. Thirdly, in the remaining three subsets with average performance, the rise in all indexes indicated that Q-Refine can identify and modify the LQ/MQ region and retain the HQ. Table \ref{tab:all} also proved in databases constructed by different T2I generation metrics with different performance, Q-Refine can provide an HQ refining result for all AIGIs.

\vspace{-1.5mm}
\subsection{Ablation Study}
\label{sec:ablation}
\begin{table}[t]
\centering
\caption{The AGIQA-3K\cite{database:agiqa-3k} refining result after abandoning different Q-Refine pipelines. The best result is noted in \textbf{\CLC{red}}.}
\label{tab:ablation}
\begin{tabular}{l|ccc}
\toprule
                Pipelines    & Overall$\uparrow$                                & Aesethic$\uparrow$                              & Fidelity$\downarrow$                               \\ \hline
(1)+(2)+(3) & {\color[HTML]{FF0000} \textbf{0.7232}} & {\color[HTML]{FF0000} \textbf{0.6021}} & {\color[HTML]{FF0000} \textbf{22.463}} \\
(1)+(2)    & 0.6604                                 & 0.5610                                 & 32.079                                 \\
(2)+(3)    & 0.6897                                 & 0.5884                                 & 24.373                                 \\
(1)+(3)    & 0.6315                                 & 0.5445                                 & 29.917                                 \\
(2)       & 0.6165                                 & 0.5147                                 & 34.299                                 \\
(3)       & 0.6852                                 & 0.5571                                 & 29.332                                 \\
\bottomrule
\end{tabular}
\end{table}

To quantify the contributions of three pipelines of Q-Refine, we abandon its stage (1)/(2)/(3) pipelines respectively in this section. As a side-effect module, (1) does not appear alone. The result in Table \ref{tab:ablation} indicates the positive effect of add-noise on subsequent denoising, as the noise from (1) greatly improves the image quality refined by (2). Both (2) and (3) have a positive effect on the refining task
, which are responsible for high-level and low-level optimization respectively.
When the two are combined, the image quality is further improved. Thus, all pipelines contribute to the final result.

\vspace{-2mm}
\section{Conclusion}
\label{sec:con}

In this study, targeting AIGI's unsatisfying quality, a quality-aware refiner is proposed. To enhance positive while avoiding negative optimization in the LQ/HQ region, IQA is innovatively introduced into the image refiner to provide guidance. Inspired by quality maps, three well-designed pipelines work collaboratively to optimize the LQ/MQ/HQ regions. Experimental data shows that Q-Refine improves the quality of AIGIs at both fidelity and aesthetic levels, which enables a better viewing experience for humans in the AIGC era.

\newpage
\bibliographystyle{IEEEbib}
\small
\bibliography{icme2023template}

\end{document}